\documentclass[11pt]{article}
\usepackage{float}
\usepackage[caption = false]{subfig}
\usepackage{acl2019}
\usepackage{times}
\usepackage{url}
\usepackage{latexsym}
\usepackage[utf8]{inputenc}
\usepackage[T1]{fontenc}
\usepackage{graphicx}
\usepackage{multirow}
\usepackage{wrapfig}
\usepackage{booktabs}
\usepackage{rotating}
\usepackage{hyperref}
\usepackage{todonotes}
\usepackage{enumitem}
\usepackage[T1]{fontenc}
\aclfinalcopy
\newcommand\palatino{\fontfamily{ppl}\selectfont}

\newcommand{\yelp}{\emph{Yelp! }}

\title{You Write Like You Eat: \\ Stylistic variation as a predictor of social stratification}
\author{Angelo Basile\\
  Symanto Research\\
  Nürnberg, Germany\\
  \small{{\tt angelo.basile@symanto.net}} \\\And
  Albert Gatt\\
  University of Malta\\
  Msida, Malta\\
  \small{{\tt albert.gatt@um.edu.mt}} \\\And
   Malvina Nissim\\
   University of Groningen\\
   Groningen, The Netherlands\\
  \small{{\tt m.nissim@rug.nl}} \\}

\date{}

\begin{document}
\maketitle
\begin{abstract}
Inspired by Labov's seminal work on stylistic variation as a function of social stratification,
we develop and compare neural models that predict a person's presumed socio-economic status, obtained through distant supervision, from their writing style on social media. The focus of our work is on identifying the most important stylistic parameters to predict socio-economic group. In particular, we show the effectiveness of morpho-syntactic features as stylistic predictors of socio-economic group, in contrast to lexical features, which are good predictors of topic.
\end{abstract}

\section{Introduction}
In 1966, linguist William Labov set out to corroborate experimentally his observation that in New York City, variation in the pronunciation of postvocalic [r] 
%the variable [r] in postvocalic position 
(as in "car", "for", "pour") is subject
%sensitive 
to \textit{social stratification} that is, that NYC people with different socio-economic backgrounds will realise that phoneme in different ways \cite{labov1966social,labov2006social}.

Avoiding artificially elicited language in favour of spontaneous language use, 
%settings of standard elicitation methods (interviews), Labov decided to collect data by observing spontaneous language use in daily life. 
Labov picked three large department stores from the top, middle, and bottom of the price/prestige range, under the assumption that customers (and salespersons) of these establishments would belong to different social strata.  "[Labov's study]  was designed to test two ideas [\ldots]: first, that the variable (r) is a social differentiator in all levels of New York City speech; and second, that \emph{casual and anonymous speech events could be used as the basis for a systematic study of language}." \cite[p. 40. Italics ours.]{labov2006social}

Inspired by Labov's work and the recent surge of interest in computational social science \citep{Cioffi-Revilla2016} and computational sociolinguistics \citep[e.g.][]{johannsen2015cross},
we set out to investigate whether and to what extent variations in \textit{writing style}, 
%expressed in terms of a variety of aspects at different level of linguistic analysis, such as   syntactic variation as well as lexical choice, is sensitive (and to what extent) to 
analysed in terms of several linguistic variables, are influenced by 
socio-economic status (\textbf{RQ1}; see below). To do so, we use user-generated restaurant reviews on social media. 
%In this respect, we care to highlight the 
User-generated content bears important similarities to Labov's "casual and anonymous speech events" on at least two fronts: 
%and the typical posts that can be written on social media: 
1) \emph{anonymity} is here still preserved since we are not including personal information about the authors; furthermore 2) social media are now recognised in the literature as a source of naturally (i.e. \emph{casual}) occurring text that can be used to investigate various sociolinguistic phenomena \cite{herdagdelen13_twitt_n_gram_corpus_with_demog_metad,Pavalanathan2015}. 

Labov's use of the \emph{prestige} of a store as a proxy for the social class of its customers and employees could be seen as a precursor of \textit{distant supervision}, an approach which we employ in this study. We leverage online restaurant reviews, and our assumption for acquiring labels is that the socio-economic group of a restaurant's patrons is in some measure predictable from its price range.
%expensive restaurants are usually visited by people belonging to higher 
%social classes and cheaper restaurants by people belonging to a lower class. The average meal price in a restaurant is our proxy for social class.
% A quick comment from an historical perspective: the \emph{results} of \citet{labov2006social} can be seen as an inspiration for our study and it would be simple to draw a line that connects his work to ours; however, we want to emphasise that our \emph{method} too is very similar to the one used by Labov. 

Using this data, we seek to address the following research questions: (a) To what extent can socio-economic status be predicted from a person's text (RQ1); (b) Can socio-economic groups be differentiated on the basis of syntactic features, compared to lexical features (RQ2)?

% want to understand if socio-economic status is associated specifically with \textit{syntactic variation} (\textbf{RQ2}) and with \textit{abstract features} extracted at the word-surface level, aimed at reducing the influence of topic (\textbf{RQ3}).  
% Summarising, here are the three main questions underlying this work:

% \begin{itemize}
% \setlength\itemsep{0em}
%     \item RQ1: can we predict the socio-economic status of a person from the text s/he writes?
%     \item RQ2: can socio-economic groups be determined based on stylistic, especially syntactic, variation?
%     \item RQ3: can we use abstract features to predict presumed socio-economic status?

% \end{itemize}

\paragraph{Contributions} Our contribution consists of 1) a silver dataset containing user-generated reviews labelled with a (distantly obtained) approximation of the socio-economic status of their author, based on the price range of restaurants; 2) a neural model of stylistic variation that can predict socio-economic status with good performance, and 
%3) a discussion of writing style in connection with socio-economic status as derived from restaurant reviews. 
3) an account of the most important features of style that are predictive of socio-economic status in this domain.
Our work can be viewed as a contemporary take on Labov's approach, with hundreds of subjects instead of only a few, and with a much larger range of proxies for socio-economic grouping, exploiting user-generated content as a natural communicative setting in which stylistic parameters can be sourced to study variation.

To favour reproducibility and future work, we make all code available at \url{https://github.com/anbasile/social-variation}.\footnote{The repository contains all code and models which can be run by acquiring the freely available Yelp dataset.}

% \todo[inline]{this we will do another time...
% validation of claim in all possible ways

% - based on literature

% - based on data analysis (any form of correlation)

% - human judgement: dataset provided to two sets of annotators (4 total) who annotate either the type of restaurant (2) or the type of person (2) plus all annotate a couple of other characteristics over which we can calculate agreement (like gender and age) 
% }

\section{Data and Labels}
To work on our questions we need user-generated texts, and a proxy 
%for acquiring the socio-economic status of the author. 
to facilitate distant labelling of an author's socio-economic status. Reviews are ideal sources of user-generated content: they are not too noisy and are of sufficient length to enable paralinguistic and stylistic parameters to be identified. 
%We decided to use reviews: their form is acceptable and not too noisy, the length is usually sufficient to allow the personality and other traits of an author to emerge, and they are easily available in a machine-readable format. In particular, we decided to use
\textit{Restaurant} reviews also carry
%, as they 
%have the advantage of carrying 
information about the restaurants themselves, especially their price range, which we can use as proxy (see below). 
%Specifically, w
We use the \href{https://www.yelp.com/dataset}{\textit{Yelp!} Dataset}: it is released twice a year from \emph{Yelp!}, a social network where users discuss and review businesses like restaurants, plumbers, bars, etc.\footnote{This data is released within the context of the \href{https://www.yelp.com/dataset/challenge}{\emph{Yelp! Challenge}}, a multi-domain shared task which has attracted attention in NLP primarily for benchmarking text classification \citep[e.g.][]{yang2016hierarchical}). We use the dataset released for Round 11.}

The review corpus contains more than 5 million documents, from over 1 million authors, with a Zipfian distribution: a small number of authors publish most of the reviews, while most of the authors only leave one review. 
%Since we are grouping reviews per author and filtering out those authors with only one review, the potential size of our final dataset goes from more than one million authors to less than a thousand.
Grouping reviews per author and filtering out authors with only one review reduces the final dataset to fewer than a thousand authors, though this set of reviews is large and allows us to 
%We can nevertheless obtain a large collection of reviews which allows us to 
\emph{infer} demographic information about the reviewers \citep[see also][]{hovy15_user}.

\paragraph{Language}
The \yelp dataset contains reviews written in multiple languages, though the vast majority are in English. We use \texttt{langid.py} \citep{Lui:2012:LOL:2390470.2390475} to automatically detect and filter out non-English instances. The need for both good parsing performance and large quantity of text limits us from working with data from other languages.

\paragraph{Price range as proxy}
 To annotate the \yelp dataset with labels which denote the social class of the authors we adopt the paradigm of \emph{distant supervision}. We take the price range of the restaurant as a proxy for socio-economic status. 
 %Before we explain this procedure, we need to put forward a caveat.
%The Yelp business data set contains information about the
The average price of a meal in a restaurant is encoded by four labels: \textdollar, \textdollar\textdollar, \textdollar\textdollar\textdollar, \textdollar\textdollar\textdollar\textdollar. As a first, coarse step, we accept this representation and divide our population into four groups.%, from the lowest to the highest socio-economic status.

\begin{figure}[h]
    \centering
    \includegraphics[width=0.9\columnwidth]{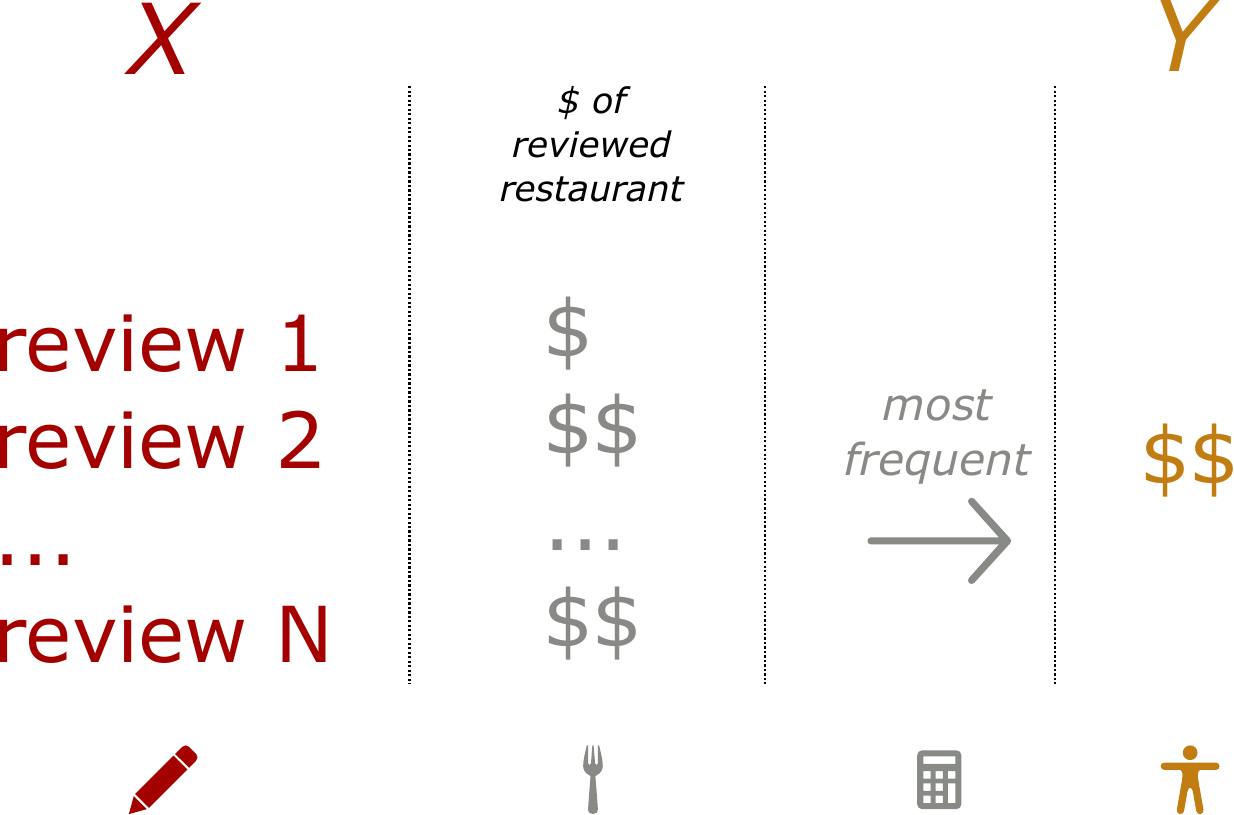}
    \caption{An illustration of the distant supervision process. Reviews from a single author are grouped together, the price range of the visited restaurants are collected and the most frequent value is assigned as label to the user. Our goal is predicting the assigned label \textit{Y} from the text \textit{X}.}
    \label{fig:drawing}
\end{figure}

We group all of the reviews per author and represent each author as a vector, where each element is the price range of a restaurant reviewed by the user. We compute the mode 
%function 
of this vector and the resulting value becomes our silver label. In short, we use the price label of a restaurant as an indicator of the socio-economic group(s) to which its patrons belong,
%cast the price label of restaurants to their visitors, under the rationale 
under the assumption that the price-range of the most visited venue will be the most indicative of the socio-economic status of a given reviewer. 
Figure \ref{fig:drawing} illustrates the process. 

\begin{figure}[h]
    \centering
    \includegraphics[scale=0.3]{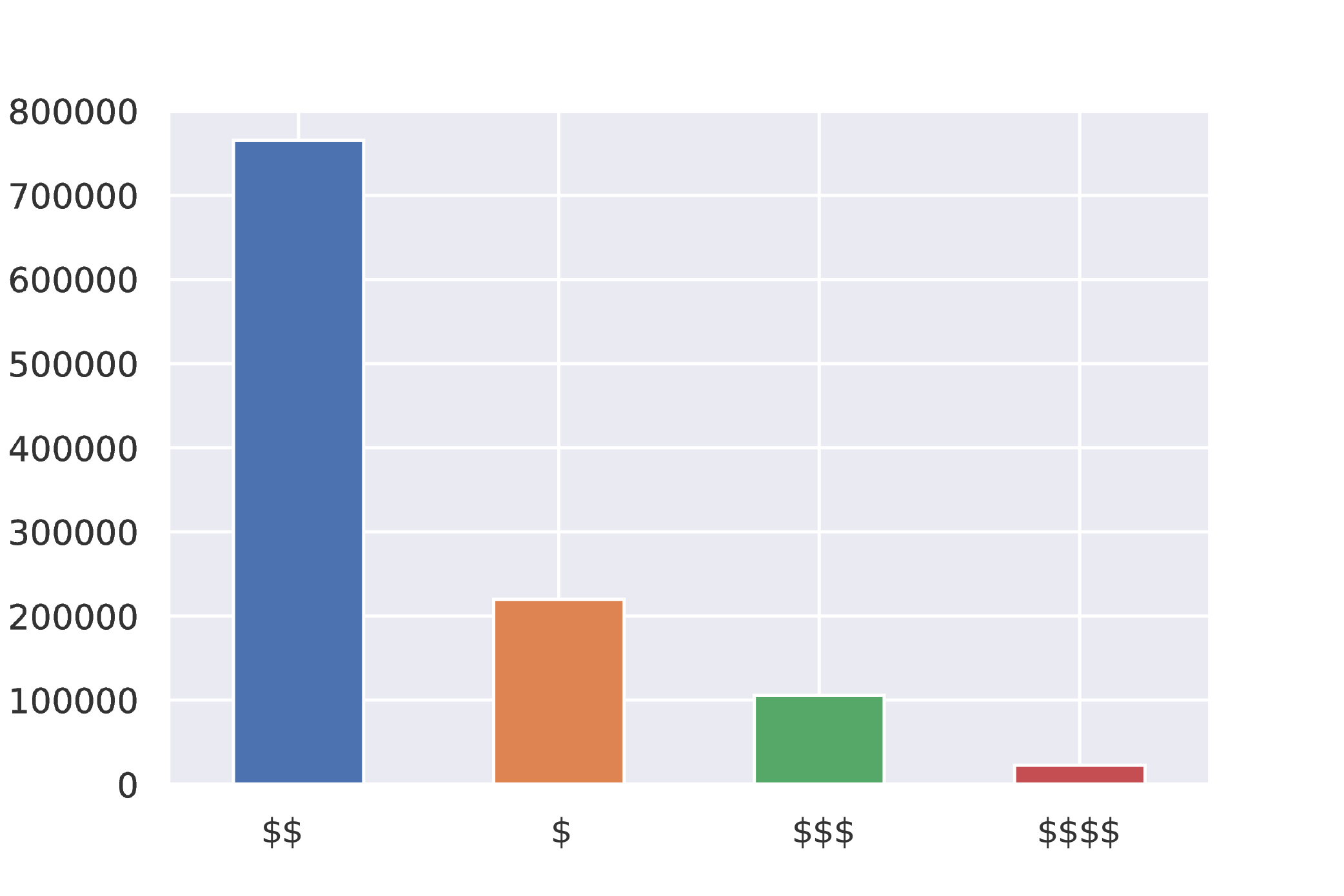}
    \caption{Author distribution before filtering. While users belonging to class \textsc{\$\$\$\$} might visit cheaper places, the same is not true in the opposite direction: this explains the small size of class \textsc{\$\$\$\$}.}
    \label{fig:distrib}
\end{figure}

This coarse representation must undergo further refinement, to satisfy three requirements:
%we set for labelling our data: 

\begin{enumerate}[label=(\alph*)]
 \item Label reliability: we want the most representative users only, that is, only those users whose restaurant price-range falls \textit{consistently} within a restricted set of categories;
    \item Sufficient textual evidence: we want as much text as possible in general, and the highest possible number of reviews per user; 
    %that visit mostly one price-range and exclude those who visit them all equally.
    \item Balance:  the raw data is highly skewed towards class \$\$ (Figure \ref{fig:distrib}), but for our experiments we want  equally represented classes to avoid any size-related effects.
    \end{enumerate}

    %Threshold to have enough textual evidence and enhance reliability (authors with one review only are not indicative label-wise, and offer too little text).
    
%    In order to 
%    First, we want to be confident that a label accurately describes an author: if an author leaves only one review, then it is impossible to infer anything about her income. 

%     Thus, we filter out the authors whose number of reviews is below a certain threshold. We set this threshold experimentally to a minimum of nine reviews, %as a way to rely
%     seeking a balance between sufficient textual evidence, maximising the number 
%     to balance between sufficient textual evidence on the one hand, and maximising the number of authors, and ensuring a balanced dataset. 
    
%     Most importantly, we also want to take care of the following problem: some authors only visit restaurants belonging to a certain price range, while others visit places belonging to all four classes.  minimise the \textit{entropy} in the label space: 
    
%     We filter out the authors who are not good representatives of their class using an entropy-based method: we describe this technique in more details in the next paragraph. 

% Lastly, we want to build a balanced data set: the raw data is highly skewed towards class \$\$ as can be seen in Figure \ref{fig:distrib}; we solve this issue by downsampling the data set to the size of the smallest class. 

\noindent In order to address (a), we employ an entropy-based strategy to filter out noisier data points. This is described below. For the size- and balance-related points (b)~and~(c), we perform two operations over the entropy-filtered dataset. First, we require a minimum number of reviews per author to ensure sufficient evidence per reviewer without excluding too many instances; we empirically set this threshold to nine reviews. Second, we downsample the larger classes to the size of the smallest class. 

\paragraph{Entropy-based refinement}
%Using silver data, the more noise we can get rid of, the better. Proxying socio-economic status with restaurant price-range, we need the `purest' reviewers, namely those that are most consistent regarding the price range of the restaurants they visit. 
Table \ref{tab:org3f61c52} shows two data points for two instances (reviewers $a$ and $b$): both consist of 16 reviews and both got assigned class 2 (i.e. \textdollar\textdollar) as a label, since 2 is the class of the restaurant that both authors visited most. However, as can be seen from the column \emph{labels}, the first reviewer visited restaurants belonging to all four classes, while the second one only visited restaurants of class 2: the second reviewer is clearly a less noisy data point. 

\begin{table}[h]
\centering
\resizebox{\columnwidth}{!}{  
\begin{tabular}{llrr}
\toprule
\textbf{user} & \textbf{labels} & \textbf{y} & \textbf{entropy}\\
\midrule
a& \{2: 5, 4: 4, 1: 3, 3: 4\}  & 2 & 1.37\\
b& \{2: 16\} & 2 & 0.00\\
\bottomrule
\end{tabular}
}
\caption[Two samples from the dataset]{\label{tab:org3f61c52}
Two equal-sized samples, both in group 2.
%from the data set, both containing 16 reviews per author and both belonging to group 2. 
The column \texttt{labels} contains the number of reviews per class.
%: for instance, in the second row (i.e. \{2:16\}), there are 16 reviews and they were all left for restaurants of price range \textdollar\textdollar.
}
\end{table}

%Since most users only left one review, they are represented with a vector of length one: in these cases the entropy will be equal to 0. If there is only one item, then there is no choice and therefore the probability of picking that item will be equal to 1 and entropy will be equal to 0. 
%The concept of ``less-noisy" can be expressed in terms of \textit{entropy}. 
\noindent To maximise the `purity' or consistency of reviews associated with each author, we compute the entropy over the label vector: the lower the entropy, the less noisy the reviewer and the more reliable the assigned label (y). In practice, we filter out the authors whose entropy score is above the mean of the whole dataset, 
%, as we deem their label not reliable enough. 
%The mean entropy is calculated over the whole dataset, 
estimated after removing authors with one review only. 

%One can see this process as a method for improving the purity of labels: by reducing the variation in the original label space, we converge toward more pure and thus more reliable inferred labels. 
%Figure \ref{fig:org911a26f} show the entropy distribution in our data set before and after removing the data points with only one review. It can clearly be seen that the first, unfiltered data is not well represented by the mean: we remember that the majority of the users only write one review. From the second plot we conclude that the mean can now be trusted for our purposes.

Table~\ref{tab:final-data-overview} shows the final label and token distribution, after filtering and downsampling. In Figure~\ref{fig:samples}, we show two sample reviews, one from class \$ and one from class \$\$\$\$.

\begin{table}[h]
    \centering
    \begin{tabular}{rcc}
    \toprule
        \textbf{class}       & \textbf{authors}   & \textbf{tokens}\\
    \midrule
        \$          & 138   &   10685\\ 
        \$\$        & 138   &   11874\\ 
        \$\$\$      & 138   &   14872\\ 
        \$\$\$\$    & 138   &   16595\\ 
    \bottomrule
    \end{tabular}
    \caption{Dataset overview after label filtering }
    \label{tab:final-data-overview}
\end{table}

\begin{figure*}
\centering
\begin{tikzpicture}
\node (table) [inner sep=0pt] {
%\begin{tabular}{p{4cm}p{4cm}p{4cm}p{4cm}}
\begin{tabular}{p{7cm}p{0cm}p{7.7cm}}
%\hline
\texttt{CLASS \$} & &
%\texttt{CLASS \$\$} & 
%\texttt{CLASS \$\$\$} & 
\texttt{CLASS \$\$\$\$}\\
\begin{footnotesize}
\palatino{So freaking good. That's all I'm gonna say. Don't believe me? Walk into the place and smell it. [\ldots] Will definitely go back.,Fresh, hand-made pepperoni rolls\ldots{}.. oh yeah. Their cheesy focattia  (did I spell that right?) is amazing. Take it home, throw it in the oven, drizzle a little EVOO on top and you're golden. Friendly people there. Parking sucks, but I'm not taking off a point for that! Their marinara is dee-lish,Super tasty!!!}
\end{footnotesize}
& &
% I'm notoriously skeptical of any restaurant that resides in some sort of strip mall.  You immediately lose 10 points off the ambiance/cool factor.  I also am reticent to eat anywhere that doesn't serve booze.  I just feel like the restaurant and I aren't on the same page and its a relationship destined for failure.Pizza A Metro exists, if nothing else, to blow up my preconceived notions.
%&
% The food was delicious, and I'm usually hesitant about Asian Fusion food. It might have converted me. We got the Tasting Menu, which was a good choice because we literally got to try everything. Usually in my experience, the nicer restaurants have smaller portions, but this was not the case. We had enough leftover food that could have fed us for another meal or so. I absolutely loved their crackling calamari salad, crispy spinach because it reminded me of the Korean seaweed snack, grilled szechuan beef, and cheesecake potsticker (dessert).
% &
\begin{footnotesize}
\palatino{Let me start off saying that 2 years ago my husband and I had a spectacular dinner at L'Atelier by Joel Robuchon and finally got the "Time" to visit Joel Robuchon.We got a limo service and a nice tour inside the mansion of Robuchon which was very memorable and the hostess escorted us to the dining area. Decore: In comparison to L'Atelier this place was much more chic and elegant. However, I still loved the idea to see all the chefs preparing and decorating my plates at L'Atelier.}
\end{footnotesize}\\
%\hline
\end{tabular}
};
\draw [rounded corners=.5em] (table.north west) rectangle (table.south east);
\end{tikzpicture}
\caption{Sample reviews for classes \$ and \$\$\$\$.\label{fig:samples}}
\end{figure*}

\section{Label validation: Readability Scores}
%As reasonable as it might be to assume that the price range of a restaurant is a good proxy for the income of its guests and therefore that our labelling strategy is sound, we can not yet assert much definitively about the \emph{social class} of an author: income is not the only component of a social group. Education level is usually another important differentiator between social groups \citep{bourdieu2013distinction}. 
%We hypothesise that given a measure of education, our groups should be ranked by this measure in the same order as they are ranked by an income measure. 
%Before turning to our prediction experiments, we seek additional validation of our distant labelling strategy of socio-economic status based on restaurant price range. 
While distant supervision allows the inference of socio-economic status with minimal manual intervention, it also makes interpretation of results challenging due to the threat of circularity involved in the process of collecting data and modelling it at the same time. 
Thus, 
%n addition to increasing label reliability through the entropy-based filtering we have described, 
We sought some external label validation that would further ensure the soundness of our labels (and thus our strategy). 

 \citet{flekova2016exploring} showed that the \textit{readability} of a text correlates with income: the higher the readability, the higher the income. This is also consistent with observations that readability correlates with educational level \citep{davenport2014readability}, which in itself plays a role in determining a person's socio-economic profile \citep{bourdieu2013distinction}.

 Assuming that our labels signal a person's income bracket, we test whether they correlate with readability scores, which would provide
 %observation allows us to test whether our labels (which we take as signalling income) correlate with readability scores. This would  provide 
 external validation of our distant labelling strategy. 

%We therefore computed rank correlations among groups identified via distant labelling (Yelp dollar signs) and readability scores of authors within each group. 
%We look for a correlation between the ranking of our groups according to an income measure (i.e. Yelp dollar signs) and according to readability scores, as these are then correlated with education \citep{davenport2014readability}. In other words, classes 1 (\textdollar), 2 (\textdollar\textdollar), 3 (\textdollar\textdollar\textdollar) and 4 (\textdollar\textdollar\textdollar\textdollar) should show (increasing or decreasing) ordered \textit{readability scores}.

We follow \citet{flekova2016exploring} and use a battery of readability metrics: Automated Readability Index, Coleman Liau Index, Dale-Chall Score, Flesch-Kincaid Ease, Gunning Fog score, Linsear Write Formula and the Lix index.\footnote{We use the implementation of these functions contained in the \emph{textstat} python library: \url{https://github.com/shivam5992/textstat}.} The metrics differ %among them 
in how they measure readability, but they all rely on features such as average number of syllables per sentence, average sentence length, or the percentage of arbitrarily defined complex words in the text.
We expect average readability to increase across groups from group 1 (\textdollar) to group 4 (\textdollar\textdollar\textdollar\textdollar) for all metrics except the 
%We compute the average readability per group and given the scores, we verify that their value increases in order from class 1 to class 4; for the 
Flesch-Reading score, where the metric's definition leads us to expect an inverse correlation  \cite{Flesch1943}.
%\todo{ref to metric?} 
%we test for inverse correlation, 
%since the metric ranges from 100.0 (5th grade) to 0.0 (college graduate). 

As shown in Table \ref{tab:org004395a}, with the exception of Linsear, the correlations go in the predicted direction: average readability score for group \textit{K} is always higher when compared to group \textit{K-1}. A Kruskal-Wallis test confirms that differences between groups are significant at \textit{p < 0.001}.
%\footnote{We test the statistical significance of the differences in readability between groups using the Kruskal-Wallis test and we find them to be significant a \textit{p < 0.001}}

\begin{table}[h]
\centering
\resizebox{\columnwidth}{!}{%
\begin{tabular}{lrrrrr}
\toprule
\textbf{Metrics} & \textbf{\textdollar} & \textbf{\textdollar\textdollar} & \textbf{\textdollar\textdollar\textdollar} & \textbf{\textdollar\textdollar\textdollar\textdollar}\\
\midrule
ARI & 6.48 & 6.52 & 6.59 & 6.91\\
Coleman-Liau & 7.58 & 7.76 & 8.07 & 8.41\\
Dale-Chall & 6.65 & 6.76 & 6.94 & 7.00\\
Flesch-Kincaid & 5.42 & 5.55 & 5.59 & 5.82\\
Flesch-Reading & 81.06 & 79.93 & 79.10 & 77.39\\
Gunning-Fog & 13.46 & 13.70 & 14.08 & 14.23\\
\textbf{Linsear} & 6.00 & 5.80 & 5.83 & 5.72\\
Lix & 30.70 & 31.39 & 31.69 & 32.71\\
\bottomrule
\end{tabular}
}
\caption[The mean readability scores per group]{\label{tab:org004395a}
The mean readability scores per group: the boldface metric is the only one whose results are not predicted by our hypothesis.}
\end{table}

%As a side note, it has also been shown that readability scores correlate with educational level \citep{davenport2014readability}, confirming the theory that education, in addition to income, plays a key role in defining someone's  socio-economic profile 
%differentiator of social groupings
%\citep{bourdieu2013distinction}.

\section{Task definition and rationale}
The prediction of socio-economic status from text can be viewed as a new dimension in the task of 
%With the goal of prediciting socio-economic status from text, we are adding a new dimension to the wider task of 
\textit{author profiling}. Due to the nature of the labels (ranging across four classes related to increasing price), this could be seen as an ordinal regression problem. However, following standard practice within the author profiling literature \cite{rangel2015overview,rangel2016overview}, especially regarding modelling age (where real values are binned into discrete classes), we treat this as a classification task. This approach results in a more conservative evaluation strategy (since at test time, a class is evaluated as either accurate or not). In an ordinal setting, one could weight classifier output by its proximity to the target class (e.g. \textdollar is closer to \textdollar\textdollar than to \textdollar\textdollar\textdollar). 
%has the advantage of
%\subsection{Settings}
%This problem can be seen as an ordinal regression as well as a classification task.
%\todo{Mention here ordinal regression? yes, double option, and mention ordinal for dollars, not necessarily for people. More conservative evaluation strategy, which is a good idea for a new task.} 
%We opt for a classification approach for the following two reasons: first, while the order concerns the restaurants, when we cast the labels onto the authors, we do not necessarily inherit order in stylistic variation \todo{but in socio-economic status yes}; second, a classification approach 
%providing a more conservative evaluation strategy, which 
Given the novelty of our task and data, where evaluation benchmarks and settings are not yet available, we deem the more conservative strategy as the most appropriate one.
%\todo{ALBERT PLEASE CHECK PREV PARAGRAPH}

%\todo{Mention again this is a new task for which we need an intrinsic definition, and some strong baseline, also to understand the soundness of label inference.}

Given a (collection of) review(s), the task is thus to predict the socio-economic status of its author, assigning one of four classes \{\textdollar,\textdollar\textdollar,\textdollar\textdollar\textdollar,\textdollar\textdollar\textdollar\textdollar\}. 
First we run a lexicon-based sparse model (the {\em lexical baseline}) which we take as a strong baseline (Section~\ref{sec:baseline}). Subsequently, we run a battery of dense models experimenting with a variety of abstractions over the lexicon (Section~\ref{sec:style}). 

Given the relative novelty of the task, we consider model performance as secondary to the broader scientific goal of identifying which features are determinants of variation as a function of socio-economic group. Thus, we focus on models that use different  features, at increasing removes from lexical or topic-based information, seeking to identify the main parameters of variation.

%\paragraph{Baselines}
%We set up two baselines. One is a simple baseline based on random class assignment. 

\section{Lexical baseline model}\label{sec:baseline}
Our baseline uses an `open vocabulary' approach \cite{schwartz2013personality}, a bag-of-word (BOW) representation of the text including all the words in the corpus, resulting in a vocabulary of 15858 items. We extract (3-6) word and character n-grams; no pre-processing is applied. We feed these features to a Logistic Regression model, which has the advantage of being highly interpretable, allowing us 
%: we use this feature 
to investigate to what extent the model relies on topic words. 
%or, more importantly, on stylistic features.

Using the Scikit-learn implementation \cite{pedregosa2011scikit}, we train the model on 80\% of the data, and test it on the remaining 20\%. 
%We compare results to a simple baseline based on random assignment.
%\subsection{Baseline model}
%\label{sec:orgb897462}
With an F1 of $0.53$, 
%As can be seen from Table~\ref{tab:baseline-results},
the performance of our lexical baseline is well above a random baseline ($F1=0.25$). %A simple model that uses a Logistic Regression and unigrams as features, scores an average accuracy of 0.47, obtained from a 10-fold cross-validation. A stronger model (see Table \ref{tab:org9ebf173}) that vectorises a segment of text at the character level and at the word level largely outperforms the simple model (61\% accuracy). A permutation test confirms that the results are significant.

\paragraph{Analysis} The scores of this simple model are most likely influenced by topic. While successful, a system assigning high weights to features strongly associated with cheap/expensive food, will limit the scope of our conclusions on stylistic variation. In other words, the features identified are more related to the restaurants themselves than to the writing characteristics of their authors. In Table~\ref{tab:org629b113} we report the most important features (words) per class.

% \begin{table}[!h]
% \centering
% \resizebox{\columnwidth}{!}{
% \begin{tabular}{rrrrr}
% \toprule
%  & precision & recall & f1-score & support\\
% \midrule
%         \$          &       0.53&      0.68&      0.59&        28\\
%         \$\$        &       0.37&      0.25&      0.30&        28\\
%         \$\$\$      &       0.67&      0.50&      0.57&        28\\
%         \$\$\$\$    &       0.58&      0.75&      0.66&        28\\
% \midrule
% avg/total &       0.54&      0.54&      0.53&       112\\
% \bottomrule
% \end{tabular}

% }
% \caption[Classification report for the baseline model]{\label{tab:baseline-results}}
% \end{table}

\begin{table}[htbp]
\centering
\begin{tabular}{llll}
\toprule
\textbf{\textdollar} & \textbf{\textdollar\textdollar} & \textbf{\textdollar\textdollar\textdollar} & \textbf{\textdollar\textdollar\textdollar\textdollar}\\
\midrule
fast & tried & at & excellent\\
kids & happy & clubs & gras\\
coffee & staff & wynn & we\\
customer & won & music & las\\
clean & put & pretty & steak\\
they & phoenix & night & tasting\\
order & find & club & foie\\
came & try & vegas & wine\\
always & place & buffet & course\\
pizza & salsa & hotel & vega\\
\bottomrule
\end{tabular}
\caption[The results of the model inspection]{\label{tab:org629b113}
The 10 most important word features per class. We omit character-level (ngram) features to facilitate interpretability.
%The model relies also character-level features, which are harder to interpret and therefore we avoided printing them here.
}
\end{table}

The output can be easily interpreted. In the least expensive class, we find words like \emph{coffee} and \emph{pizza}. The second class is noisier, as the model appears to capture  aspects of the reviews related to service rather than food. The two most expensive classes confirm our hypothesis since we find words like \emph{Vegas}, \emph{Wynn} (a casino in Las Vegas, USA), \emph{[foie-?]-gras}, \emph{wine} and \emph{steak}.
%; second, these two classes could probably be merged into one. Figure \ref{fig:org481e134} shows that indeed the model confuses the third and the fourth class.\todo{confusion matrix in additional materials?}
%\footnote{From a certain perspective, we can say that the model is learning how to properly classify the extreme cases: one reason for this might be that the extreme classes are more different from each other than the other two classes. We hypothesised this when investigating latent variables in Chapter 3.4: the plots of the data transformed via a dimensionality reduction technique seems to support this hypothesis.}.

What we observe from this feature analysis is that by relying on words we are capturing aspects of restaurants, to the detriment of a properly stylistic account, whose features would be more author- than topic-oriented.
%and we do not know much about stylistic aspects. In order to capture style more than topic, and thus model more the authors themselves rather than the restaurants, we need 
Capturing author-related stylstic features requires an abstraction
%to abstract 
away from the lexicon (though not necesssarily from non-content based featues of the lexicon, such as word length or structure). This might yield lower performance, but our main goal is to understand the role played by morpho-syntactic and other non-lexical dimensions of social variation, rather than achieving the highest possible score in classifying reviews.

\section{Capturing Style}
\label{sec:style}

Style and variation can be found at different levels of linguistic abstraction \citep{eckert2001style}.  
We experiment with a selection of features carefully tailored to capture different aspects of the phenomenon; each feature serves as a representation to be fed to a classifier.

First, we preserve the surface structure but get rid of most lexical information, using the bleaching approach proposed by \citet{goot18:bleac_text} (Section~\ref{sec:bleaching}). Second, we remove words and replace them with POS tags, so as to cancel out topic information entirely (Section~\ref{sec:morpho-syntax}). In the final representation, we use dependency trees and expand the POS tags into triplets to investigate syntactic variation (Section~\ref{sec:dependency}).

%\paragraph{Dense Models}

In order to properly model the structural information encoded in these non-lexical feature representations, we use a Convolutional Neural Network (CNN) classifier \citep{lecun1995convolutional}, rather than rely on sparse models as we did for our lexical baseline.\footnote{Although the aim of this paper is not a comparison between sparse and dense models over different representations, we provide all scores for all models in the appendix.}
The model consists of a single convolutional layer coupled with a sum-pooling operation; a Multi-Layer Perceptron on top improves discrimination performance between classes. We use the Adam optimizer \citep{kingma2014adam} with a fixed learning rate (0.001) and L2 regularization \citep{ng2004feature}; a dropout layer (0.2) \citep{srivastava2014dropout} helps to prevent overfitting. For the implementation we rely on spaCy \citep{honnibal-johnson:2015:EMNLP}.

\subsection{Bleached representation}
\label{sec:bleaching}

 Recently, \citet{goot18:bleac_text} introduced a language-independent representation termed \textit{bleaching} for capturing gender differences in writing style, while
abstracting away from lexical information.
%bypassing the lexicon. 
%Indeed, the principle of 
Bleaching preserves 
%is preserving 
surface information while obfuscating lexical content. 
%as much as possible the lexicon, 
This allows a focus on lexical variation as a function of personal style, while reducing the possible influence of topic as a determining factor.
%(which is indicative, of course, but trumps other stylistic aspects which instead we're interested in).\todo{just left this in brackets as a note, we need to say it somewhere.} 

We experiment with this idea under the assumption that authors belonging to different groups will show a difference in the formality of their writing, and that a bleached representation is well suited for capturing such a difference. 

In particular, we hypothesise that some of our target classes are typified by certain writing styles which differ in their formality and the extent to which they approach informal speech. Thus, we aim to capture the difference between a plainer writing style, with few or no interjections, without abbreviations and/or emojis; and a writing style which more closely approximates speech, making substantial
 %large 
 use of exclamation marks and emojis for emphasis, abbreviations, possibly incorrect spelling of words to approximate phonetic form and broad use of direct speech. 
 
As an example, the following is a list of sentences taken from different classes of our dataset:

\begin{description}
\item[\textdollar] -- \emph{hand-made pepperoni rolls\ldots{}.. oh yeah}

\item[\textdollar\textdollar] -- \emph{Their marinara is dee-lish,Super tasty!!!}

\item[\textdollar\textdollar\textdollar] -- \emph{When Jet first opened, I loved the place.}

\item[\textdollar\textdollar\textdollar\textdollar] -- \emph{compared to pierre gagnaire in paris, the food here is way less ambitious}
\end{description}

\noindent  We note that orthography seems to differ significantly between these samples: the first two would more likely be viewed as typical web texts, while the last two show a more considered or premeditated writing style.

\begin{table}[h!]
\centering
\begin{tabular}{ll}
\toprule
\textbf{token} & \textbf{bleached representation}\\
\midrule
I       & \texttt{X\_01\_True\_V\_2117}\\
really  & \texttt{xxxxxx\_06\_True\_CCVVCC\_81}\\
love    & \texttt{xxxx\_06\_True\_CVCVCC\_15}\\
pizza   & \texttt{xxxxx\_04\_True\_CCVC\_617}\\
!       & \texttt{!\_01\_False\_!\_21}\\
\bottomrule
\end{tabular}
\caption{\label{tab:bleached-representation}
An example of how a sentence is rendered by the bleached representation.}
\end{table}

Table \ref{tab:bleached-representation} shows some examples of the bleached representation under the abstraction we chose to experiment with, which are as follows. First, we extract the surface form of a word and render each character as either \emph{X} or \emph{x}, depending on whether it is capitalised or not. Second, we extract the length of each word prefixed with a {\tt 0}
%while taking care to prepend a {\tt 0} to it 
to avoid confusion with the frequency of the word (indicated by the number at the end of the bleached string). A boolean label signals whether the token is alphanumeric or not: this feature can be informative in capturing, for instance, the use of emojis. Finally, we approximate the original surface form by substituting all the English vowels with the letter \emph{V} and all the English consonants with the letter \emph{C}.

\subsection{Morpho-syntax}
\label{sec:morpho-syntax}
%Our first experiment for investigating syntactic variation between social classes consists in POS-tagging each word and trying to profile authors using this transformed version of the corpus. If such an approach would work, we would both limit the effect of topic in our model (a phenomenon which is probably appearing in the baseline experiment) and prove our hypothesis. We use the Universal POS Tagset \cite{petrov2012universal}.
As a more definitive move away from lexical information, we label each word by its POS-tag, using spaCy \citep{honnibal-johnson:2015:EMNLP} and the universal tagset \citep{petrov2012universal}. Within this experiment, we train our model using only such a representation, thus inhibiting topic-related features from becoming prominent. We assume that a good performance of the classifier under such conditions %translates into a good argument for 
provides support for the existence of phenomena related to social variation at the morpho-syntactic level.

\subsection{Dependency trees}
\label{sec:dependency}
Previous research on stylistic variation as a function of age and income shows an important difference in syntax use between groups \citep{flekova2016exploring}. However, this work reports results based on a shallow interpretation of syntax, i.e. the authors measure the ratio of POS tags in the text: such a strategy is dictated by the relatively poor performance of parsers on the domain investigated by \citet{flekova2016exploring}, i.e. Twitter. \yelp reviews are closer to canonical English, which allows us to obtain a full syntactic analysis of each document, adopting a strategy closer to that of \citet{johannsen2015cross}.
%: this strategy follows the work of \citet{johannsen2015cross} on variation over age and gender.

\begin{figure}[h]
    \centering
    \includegraphics[width=.9\columnwidth]{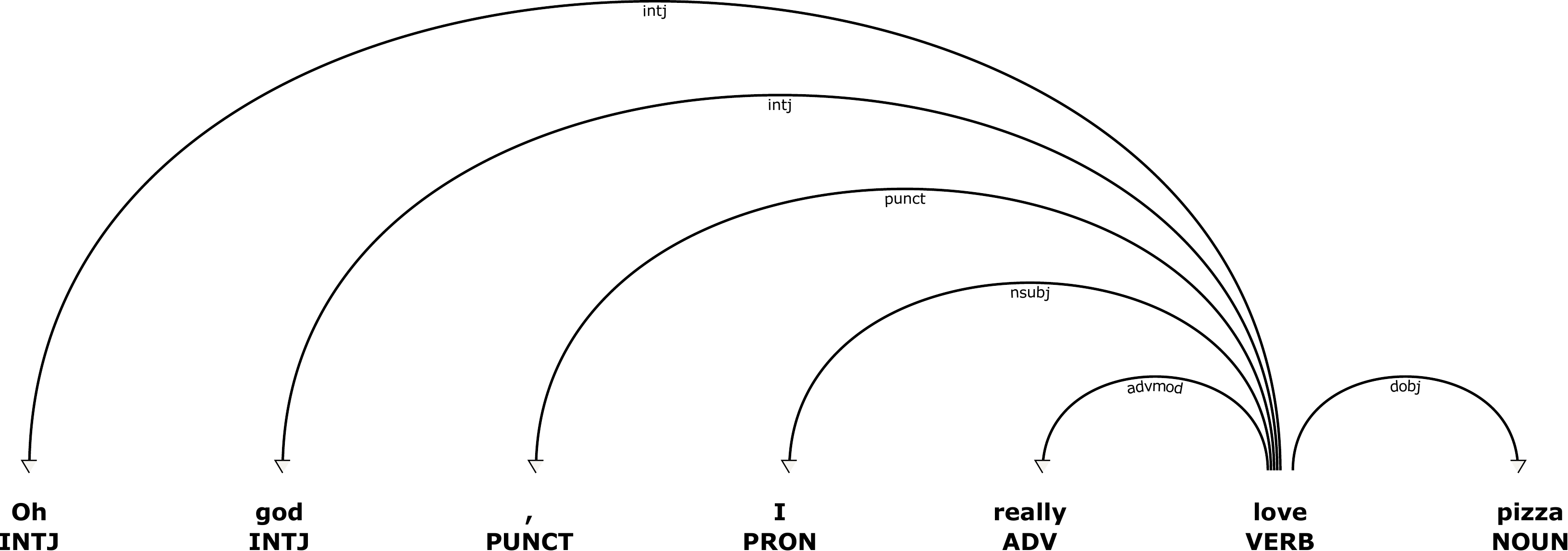}
    \caption{An example of a parsed sentence using Universal Dependencies.}
    \label{fig:sample-parse}
\end{figure}

We first parse our corpus using a pre-trained dependency parser, namely \citet{honnibal-johnson:2015:EMNLP}'s parser\footnote{We use the largest pre-trained available model, \texttt{en\_core\_web\_lg}.}, which achieves state-of-the-art accuracy on English. Figure \ref{fig:sample-parse} shows an example. % of a sentence's parse. 

\begin{figure}[htb]
    \centering
    \includegraphics[width=1.0\columnwidth]{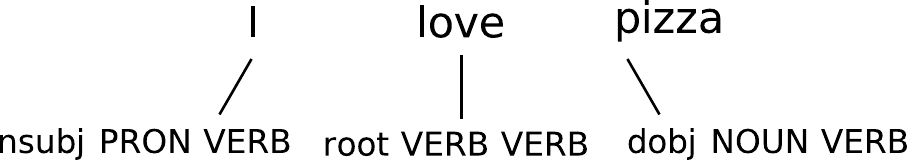}
    \caption{An example of the syntactic feature representation.}
    \label{fig:sample-syntax}
\end{figure}

We then transform each word into a triplet that consists of: 1) the POS tag of the word, 2) the incoming arc  and 3) the POS tag of the head, as shown in Figure \ref{fig:sample-syntax}.
%shows its derived representation, which
This is fed as feature to the classifier. \citet{johannsen2015cross} use a `bag-of-relations' representation in combination with a ${\chi}^2$ test, discarding some structural information in order to ease comparison across languages: here, we rely on the performance of a sequence model (i.e. the CNN classifier) over the transformed dependency tree. As we do in Section \ref{sec:morpho-syntax}, we assume that a good performance of the classifier points toward the existence of significant syntactic patterns between groups. 

\section{Evaluation}

\begin{figure*}
    \centering
    \subfloat[bleaching]{\includegraphics[width = 0.33\textwidth]{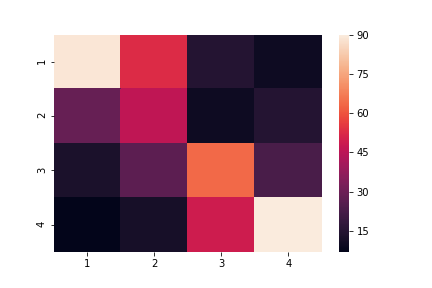}}
    \subfloat[POS]{\includegraphics[width = 0.33\textwidth]{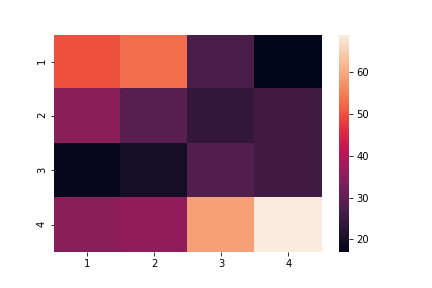}}
    \subfloat[dependency trees]{\includegraphics[width = 0.33\textwidth]{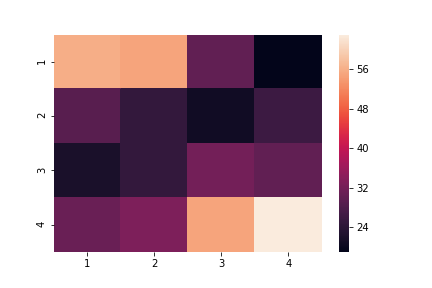}}
    \caption{Confusion matrices for the CNN models using bleached representations, POS, and dependency trees.}
    \label{fig:CM}
\end{figure*}

%Given the relative novelty of the task, our scientific goal is to assess the role of morpho-syntactic and other non-lexical features in determining variation as a function of socio-economic group, more than aiming at high performance. 

We focus on the comparison of several models against one another and especially against the lexical baseline. This will let us single out which features, or which levels of abstraction (see Section~\ref{sec:style}), best model style when topic information is reduced or eliminated. For completeness, we also report on the results obtained by a CNN-based version of the LR lexical baseline from Section~\ref{sec:baseline}.

In Table~\ref{tab:main-results}, we report results training our models on 80\% of the data and testing them on the remaining 20\%, using exactly the same split as for the simple lexical and random models (Section~\ref{sec:baseline}). Note that the results are averaged over two runs: we ran the CNN twice for each representation, since it is known that multiple runs of the same neural model on the same dataset can yield significantly different results due to underlying random processes \citep{reimers2017reporting}.
%Our main goal is to understand the role played by morpho-syntactic and other non-lexical aspects in social-variation, more than achieving the highest possible score in classifying reviews: for this reason, in Table \ref{tab:main-results} we report f1-score of several models and compare them against our two baselines (random and lexical).\todo{We still need to add settings (split)} In the following sections we describe and discuss the performance of the various models.

%SETTINGS: mention split; mention standard measures.

\begin{table}[h]
    \centering
    \begin{tabular}{lc}
    \toprule
         \textbf{model} & \textbf{F1}\\
    \midrule
         random baseline        &   0.25\\
         LR BOW (lexical) baseline        &   0.53\\
    \midrule \midrule
         CNN lexical            &   0.54\\
         \midrule
         CNN pos tags           &   0.33\\
         CNN dependency tree    &   0.52\\
         CNN bleaching          &   0.46\\
    \bottomrule
    \end{tabular}
    \caption{F1-scores of the Logistic Regression (LR) and Convolutional Network (CNN) models on our dataset.}
    \label{tab:main-results}
\end{table}

As a general comment, from a class perspective, we  observe that class 4 is the easiest to model, while class 2 is the most difficult, for all CNN models (see the confusion matrices in Figure~\ref{fig:CM}). This complements the observation made earlier in relation to Table \ref{tab:org629b113}, where it was noted that class 2 is also noisier at the lexical level.

\paragraph{Lexical} This model serves as a comparison to the LR-based lexical baseline model, while also providing a CNN-based version of this model to ensure fair comparison of a lexical or topic-based strategy against other, non-lexical, CNN models.
%we ran mainly for comparison, in two ways: for comparison to the sparse model, to see if a lexicon-based CNN model was competitive (it has been shown that simple sparse models perform best for author profiling)\todo{fix}; and for comparison to the models based on lexically-poor representations.
The lexical CNN achieves approximately the same results as the LR-based lexical baseline, with an overall F-score of 0.54.

\paragraph{Bleaching}
Our CNN model trained on bleached representations  shows the lowest performance, though still above random baseline.\footnote{When this feature is used in the logistic regressor instead, it shows good performance. See the Appendix for details.}
%\todo{still need to add some explanation}
%The model is outperformed by the sparse ones: we believe that this is due to the limited amount of data available. As can be seen from Table \ref{tab:abstract-results} the model scores on average 46\% f1-score.
This suggests that abstract, word-level features do have some predictive value, but they do not capture enough lexical content to surpass a simple lexical model that classifies based on topic-based features. At the same time, this result also indicates that the {\em shape} of the lexical items used by authors (the outcome of bleaching) is a less reliable predictor of socio-economic status than certain morpho-syntactic properties.

\paragraph{POS tags}
When using only POS information without words, we find that, as can be expected, performance drops ($F=0.33$). 
%The permutation test shown in Figure \ref{fig:orge2c1905} shows that although the cross-validated result is above the baseline, it is not significant. 
From the confusion matrix reported in Figure~\ref{fig:CM}, it appears once again that class~2 is the hardest class to predict. 
% \begin{figure}[h]
% \centering
% \includegraphics[width=.7\linewidth]{images/morpho-permutation-test.eps}
% \caption{\label{fig:orge2c1905}
% Permutation test for the morphological model.}
% \end{figure}

\paragraph{Dependency Trees}
As an abstraction strategy, this works best out of the three we have tried, and is competitive with the neural lexical model and the logistic regressor. As Figure~\ref{fig:CM} shows, the model is also predicting each of the four classes more consistently than the other two models.
This suggests that we are able to leverage syntactic information as a  predictor of social variation, echoing the findings of \citet{johannsen2015cross} in a different sociolinguistic domain. Higher accuracy is also achieved without any topic bias, thus providing better evidence that we moved away from a model that predicts which restaurants are the topic of discussion, and moved closer to an account of authorial style.
%restaurant descriptions per se, and got closer to model style.
%\todo{FIX!}

%  The results of our syntactic model show that a BOW approach is not optimal for handling the structured information that is encoded in syntax. The BOW model achieves a significant positive results, which outperforms the POS only model. However, the CNN outperforms the BOW model by almost twenty points in f-1 score. The sparse model achieves 31 \% in accuracy, while the dense one achieves an accuracy of 49,5\% averaged over two runs. These two results are not really comparable since the sparse model is cross-validated while for the neural one we rely on a data split. The sparse model trained on the same split reaches 46\% in accuracy.  

% As explained in \ref{sec:orgfaf00c6}, we developed in parallel two different models, a sparse and a dense one, to compare their results and for obtaining insights on the actual data being modelled. From the results reported in Table \ref{tab:syntax-results}, we conclude that the dense model significantly outperforms the BOW approach. 

% One reason why we decided to develop a sparse model was our need to inspect the model in order to understand what features are being used: in Section \ref{sec:orgb897462} we were able to do this and we could spot a problem in our baseline model. However, here the data structure that we are modelling does not allow for this analysis to be performed, since the structural information encoded in a syntactic tree is being broken during the vectorization process. This analysis will probably benefit from an attention model.

\medskip
\noindent We believe these results provide a positive answer to our main research question (\textbf{RQ1}): to the extent that authors can be distantly grouped according to their socio-economic status, it is possible to differentiate among them on the basis of stylistic parameters. As for our other question, we find that the two strongest predictors of our labels are lexical information on the one hand, and syntactic dependencies on the other. We attribute this to the fact that these models are ultimately classifying different things: a lexically-based model relies on topic and thus predicts the type of restaurant. A syntax-based model is a better approximation to individual style. That these two models achieve very similar F1 scores (0.52 vs 0.54) can be attributed to the fact that filtering and downsampling created a more consistent dataset in which authors were consistently grouped in specific restaurant price ranges. These two models show that it is possible to differentiate among the resulting classes both on the basis of type of establishment (the lexical model) and on the basis of stylistic features in the writing style of its patrons (the syntactic model).
% between social classes and can it be used for prediction purposes?). As far the other two research questions are concerned... \todo{FIX}

%We also note that sparse models are not always easier to understand when compared to dense ones.

% \begin{table}[!h]
%     \centering
%     \begin{tabular}{lrrlrlr}\toprule
%     model & class & precision & recall & f1-score\\
%     \toprule
%     \multirow{5}{*}{Sparse} & 1 & 0.35 & 0.41 & 0.38\\
%      & 2 & 0.25 & 0.18 & 0.21\\
%      & 3 & 0.29 & 0.23 & 0.26\\
%      & 4 & 0.35 & 0.46 & 0.39\\
%      & avg/total & 0.31 & 0.32 & 0.31\\ \midrule
%     \multirow{5}{*}{Dense} & 1 & 0.54 & 0.55 & 0.54\\
%      & 2 & 0.40 & 0.40 & 0.40\\
%      & 3 & 0.41 & 0.57 & 0.47\\
%      & 4 & 0.70 & 0.51 & 0.57\\
%      & avg/total & \textbf{0.54} & \textbf{0.47} & \textbf{0.50}\\ \bottomrule
%     \end{tabular}

%     \caption{Results for the both the sparse and dense models using syntactic features. The results reported for the dense model are averaged across two runs.}
%     \label{tab:syntax-results}
% \end{table}

\section{Related Work}
\label{sec:related}

The idea that socio-economic status influences language use and is a determinant of language variation has been central to sociolinguistic theory for a long time \citep{Bernstein1960,Labov1972,labov2006social}. Labov's work could be viewed as an early form of distant supervision, exploiting established categories (e.g. the price and status of establishments such as department stores) to draw inferences about variables related to social stratification. The work presented here takes inspiration from this paradigm, and contributes to the growing literature on distant supervision in NLP \citep{Read:2005:UER:1628960.1628969}, especially in social media \citep[e.g.][{\em inter alia}]{plank2014adapting,pool2016distant,fang2016learning,basile2017predicting,Klinger2017}.

Computational work on style -- i.e. linguistic features characteristic of an individual or group \cite{Biber1988} -- has focussed on demographic or personal variables, ranging from geographical location and dialect \cite{zampieri2014report,han2014text,Eisenstein2013a} to age and gender \cite{Argamon2007,Newman2008,sarawgi2011gender,johannsen2015cross,Hovy2015}, as well as personality \cite{Argamon2005,verhoeven2016twisty,youyou2015computer}. An general overview of computational sociolinguistics can be found in \citet{Nguyen2016ComputationalSA}.
%For example, variation has been successfully leveraged to identify the geographical provenance of speakers or writers \cite{zampieri2014report,han2014text} and the way phonological features of dialect influence writing style \citep{Eisenstein2013}. Individual factors such as age and gender have also been classified on the basis of textual features with varying degrees of success \citep[e.g.][]{Argamon2007,Newman2008,sarawgi2011gender,johannsen2015cross}, with some recent demonstrations that these two parameters also influence the performance of standard NLP tools \citep{Hovy2015}. Language variation is at work also between groups of individuals with different psychological traits: \citet{verhoeven2016twisty} show that personality is correlated with language use and that it can be predicted from text with a relatively high accuracy. Apart from text, non-linguistic features can also be exploited by algorithms to accurately predict personality types: such systems have been shown to even outperform human judgements \citet{youyou2015computer}. 

By contrast, there has been relatively little work on socio-economic status. \citet{flekova2016exploring} show that textual features can predict income, demonstrating a relationship between this and age. \citet{lampos2016inferring} also report good results on inferring the socio-economic status of social media users from text. Like the present work, they use distant supervision, exploiting occupation information in Twitter profiles. 
Our work differs from these precedents in that 
%we seek to validate our labels both automatically and via human validation and 
we investigate a broader range of lexical, morphological and syntactic features in a novel domain.

Previous work specifically on the language of food has also found that social media data can be used to validate sociological hypotheses, such as the importance of a specific meal in a certain geographical region \cite{fried2014analyzing}. Somewhat closer to the present work, \citet{jurafsky2014language} finds an interesting correlation between the price range of a restaurant and the lengths of food names on its menu.

%FIRST VERSION - 25/05/19
%On the language of food specifically, \citet{jurafsky2014language} finds an interesting correlation between the price range of a restaurant and the lengths of food names of its menu; \citet{fried2014analyzing} shows that tweets about food can be used to infer sociological information such as the importance of a specific meal in a certain geographical region.

%%Related Work on the languae of food
% Missing References
% ---------------------------------------------------------------------------
% Although Labov has done quite some research on sociolinguistic variation, the sociolinguistic references could be more up to date. The authors are kindly reminded to follow/read papers from these journals (e.g. Language, Variation & Change, Language in Society). Research presented in these journals will inspire the authors for new RQs and also to support/explain their findings. 
% 
% The Language of Food (Dan Jurafsky): There is a section about the link between the lengths of food names on the menu & price range of the restaurants. 
% 
% Nguyen, D, et al. (2016). Computational Sociolinguistics: A Survey, Computational Linguistics, 42, 537-593 (caveats of linking linguistic style with social factors & social media)
% 
% The authors could also mention about the user biases in restaurant reviews (e.g. https://arxiv.org/pdf/1604.00417.pdf)
% This bias could influence their style as well. 
% 
% Fried, D. et al (2014) Analyzing the Language of Food on Social Media. 
% https://arxiv.org/pdf/1409.2195.pdf
% 

\section{Conclusion}
Inspired by Labov
%seminal work on social stratification, 
and encouraged by recent interest in computational sociolinguistics, 
%the recent surge of work on computational sociolinguistics, 
we developed accurate neural models to predict socio-economic status from text. While lexical information is highly predictive, it is restricted to topic. In contrast, syntactic information is almost as predictive and is a much better signal for stylistic variation.

From a methodological point of view, we can draw two conclusions from this work. First, as has been noted \citep{plank2016multilingual}, neural networks can perform well with relatively small datasets, in this case proving competitive with the sparse models that are usually favoured in author profiling \cite{malmasi2017report,basile2018simply}.
%univ students (highly educated but poor).Labov must have had the same, though. Accepted noise?
% Also, as it has been recently observed \citep{plank2016multilingual}, we confirm that neural networks do not always require large amounts of data to perform well. And at least for some aspect of author profiling, they are definitely competitive with sparse models, usually deemed more performant on this task \cite{malmasi2017report,basile2018simply}.
Second, distant supervision with proxy labels for socio-economic status yields useful insights and is validated externally via readability scores. This is encouraging for
% Our strategy of proxying labels representing socio-economic status through restaurant price ranges was consolidated by results themselves, and especially by external validation through readability scores. This is good news for 
further studies in computational social science in ecologically valid and relatively labour-free settings.
% , especially through the ecological and labour-free paradigm of distant supervision. 

Nevertheless, there are limitations of distant labelling and social media data --- with issues related specifically to the language of food \cite{Askalidis2016UnderstandingAO} --- that we will take into account in future work. First, we wish to investigate the role of additional variables (such as age and gender). Second, we will take steps to mitigate the risk of fake reviews and validate the distant labelling with human annotation.
%ORIGINAL VERSION - as submitted
%Nevertheless, there are limitations of distant labelling and social media data that we will take into account in future work. First, we wish to investigate the role of additional variables (such as age and gender). Second, we will take steps to mitigate the risk of fake reviews and validate the distant labelling with human annotation.

% we need to bear in mind the limitations of distant labelling and the interpretation we make of those labels. We are accounting for one variable only, while others might be at play, such as age, education\todo{fix}, gender, etc. Also, reviews might not be all genuine\todo{any ref to fake reviews?}, but this is an aspect that all work on social media must take into account\todo{add something better!}
% In future work, we are also seeking to validate the outcomes of our distant labelling strategy with human annotators.

\section*{Acknowledgements}

We would like to thank the three anonymous reviewers who helped us improve the quality of this paper. The first author's contribution was made while at the Universities of Malta and Groningen as part of the Erasmus Mundus M.Sc. Program in Human Language Science and Technology.

\bibliography{references,references_extra}
\bibliographystyle{acl_natbib}

\section*{Appendix: Additional results}

\begin{table}[h]
    \centering
    \resizebox{\columnwidth}{!}{%
    \begin{tabular}{l|rrrrr} \toprule
     model & class & precision & recall & f1-score\\ \toprule
     
    \multirow{5}{*}{lexical} 
        & \$          &       0.53&      0.68&      0.59\\ 
        & \$\$        &       0.37&      0.25&      0.30\\
        & \$\$\$      &       0.67&      0.50&      0.57\\
        & \$\$\$\$    &       0.58&      0.75&      0.66\\
        & avg/total   &       0.54&      0.54&      0.53\\
    \midrule
    
    \multirow{5}{*}{abstract} 
        & \$        &       0.61&      0.71&      0.66\\ 
        & \$\$      &       0.50&      0.32&      0.39\\
        & \$\$\$    &       0.39&      0.32&      0.35\\        
        & \$\$\$\$  &       0.42&      0.57&      0.48\\        
        & avg/total &       0.48&      0.48&      0.47\\         

    \midrule
    
     \multirow{5}{*}{POS-tags} 
        & \$        &       0.27&      0.43&      0.33\\        
        & \$\$      &       0.15&      0.07&      0.10\\        
        & \$\$\$    &       0.29&      0.14&      0.19\\        
        & \$\$\$\$  &       0.40&      0.57&      0.47\\        
        & avg/total &       0.28&      0.30&      0.27\\       
    \midrule
    
     \multirow{5}{*}{dependency triplets} 
        & \$        &       0.43&      0.36&      0.39\\        
        & \$\$      &       0.23&      0.21&      0.22\\        
        & \$\$\$    &       0.21&      0.25&      0.23\\        
        & \$\$\$\$  &       0.37&      0.39&      0.38\\        
        & avg/total &       0.31&      0.30&      0.31\\       
    \bottomrule
    \end{tabular}
    }
    \caption{Classification report for the sparse model using the different representations.}
    \label{tab:abstract-results}
\end{table}
\end{document}